\documentclass{INTERSPEECH2023}

\interspeechcameraready 

\usepackage[dvipsnames]{xcolor}
\usepackage{caption}
\usepackage{subcaption}
\usepackage{hyperref}
\usepackage{textcomp}
\usepackage{tikz}
\usepackage{pgfplots}
\usepackage[bottom]{footmisc}
\usepackage{bbm}
\usepackage{amssymb}
\usepackage{soul}
\usepackage{booktabs}
\usepackage{contour}
\usepackage{algorithm}
\usepackage{algorithmic}
\usepackage{subcaption}
\usepackage{amsfonts}
\usepackage{multirow}
\usepackage{multicol}
\usepackage{mathabx}
\usepackage{colortbl}
\usepackage{hyperref}

\definecolor{Gray}{gray}{0.97}
\definecolor{Grey}{gray}{0.92}
\definecolor{Graey}{gray}{0.88}
\newcommand{\SeqAug}{$\mathrm{SeqAug} $}

\title{
SeqAug: Sequential Feature Resampling as a modality agnostic augmentation method
}
\name{Efthymios Georgiou$^{1,2}$, Alexandros Potamianos$^2$}
\address{
  $^1$ School of ECE, National Technical University of Athens,  Athens, Greece \\ 
$^2$ Institute for Language and Speech Processing, Athena Research Center, Athens, Greece}
\email{efthygeo@mail.ntua.gr, potam@central.ntua.gr}

\begin{document}

\maketitle
 
\begin{abstract}
Data augmentation is a prevalent technique for improving performance in various machine learning applications. We propose SeqAug, a modality-agnostic augmentation method that is tailored towards sequences of extracted features. The core idea of SeqAug is to augment the sequence by resampling from the underlying feature distribution. Resampling is performed by randomly selecting feature dimensions and permuting them along the temporal axis. Experiments on CMU-MOSEI verify that SeqAug is modality agnostic; it can be successfully applied to a single modality or multiple modalities.
We further verify its compatibility with both recurrent and transformer architectures, and also demonstrate comparable to state-of-the-art results. 

\end{abstract}

\section{Introduction} \label{sec:intro}



Data augmentation (DA) is a powerful arsenal of techniques employed to increase the amount and diversity of raw data. When applied during training of deep neural networks (DNNs), DA offers a compelling advantage by enhancing performance and closing the generalization gap. Nevertheless, crafting augmentation techniques requires a thorough understanding of domain-specific knowledge and expertise~\cite{DA_lecun, backtranslation}. Consequently, each DA method is a tool that is attached to specific data sources, configurations, and network architectures. 



DNNs often have to process extracted features instead of raw data representations. 
Training DNNs using extracted features is motivated by the reduced computational cost compared to processing raw information, e.g., video streams, as well as the benefits of exploiting hand-crafted or learned representations. For instance, in text-related tasks, one can use large pretrained models such as BERT~\cite{kenton2019bert} embeddings. For audio tasks, hand crafted features such as MFCCs have been employed. This adds a layer of complication for DA algorithms, since standard modality specific DA techniques might not be directly applicable on feature representations.

With the benefits of data augmentation techniques established in other domains, a crucial research question arises: \textit{Can we develop a data augmentation technique that is well-suited for feature learning setups?} Such a technique should have three key properties: 1) boost performance in a similar manner to traditional data augmentation methods, 2) it should be modality-agnostic, enabling it to work with a broad range of heterogeneous modalities, and 3) work seamlessly for various DNN/ML model architectures. Recognizing the potential advantages of a feature-based data augmentation technique, we focus our research on enhancing multimodal sequential features, such as BERT embeddings, and aligned MFCCs per token.

\begin{figure}[!htb]
    \centering
    \includegraphics[width=0.90\linewidth]{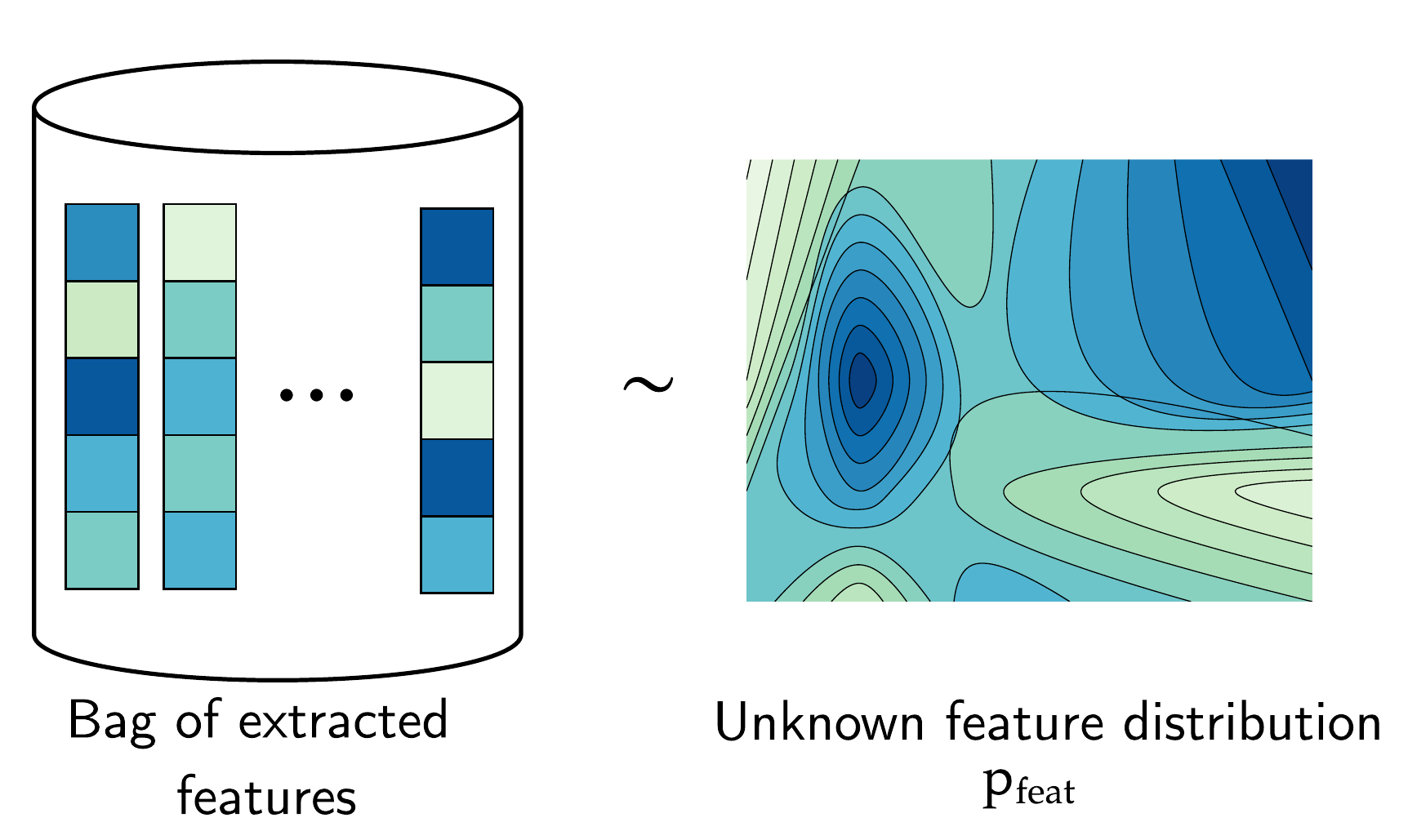}
    \caption{The bag of all extracted features. We assume that sampling the bag of features approximates the unknown feature disrtibution $\mathsf{p_{feat}}$.}
    \label{fig:p_feat}
\end{figure}


In this work \footnote{Code to be made available upon acceptance}, we propose $\mathrm{SeqAug}$, a modality agnostic augmentation method that operates on sequences of features.
The fundamental concept involves augmenting a sequence by resampling feature values from the underlying feature distribution. We make the assumption that each feature vector in the sequence may be drawn from an unknown feature distribution (see Fig.~\ref{fig:p_feat}).  
The resampling process is achieved by randomly selecting feature dimensions and permuting them along the temporal axis, thereby replacing the original values with existing features from the same sequence. 
This way we achieve sequential augmentations which 1) can be modeled as resampling from the original distribution and 2) do not affect the underlying semantics since resampling occurs within the sequence itself. 

We present experiments in a multimodal sentiment analysis dataset for which we have extracted features for every stream. Unlike other DA methods our approach is modality agnostic and is thus suitable for all streams of features, as well as multimodal setups. We apply $\mathrm{SeqAug}$ in each stream independently and perform the following two steps. First, we randomly pick some feature addresses, i.e., dimensions, and we then permute these addresses along the time-axis with a common permutation. Our results show that applying $\mathrm{SeqAug}$ on each stream independently improves performance. Further we find that augmenting all modalities simultaneously provides an even greater performance boost. Finally,  we show that $\mathrm{SeqAug}$ is 
able to match the baseline model performance with $20\%$ less data. Notably, our approach is simple and computationally efficient as it acts directly on the sequence features themselves.

We summarize our \textit{contributions}:
\begin{enumerate}
    \item We propose $\mathrm{SeqAug}$, a novel sequential feature resampling augmentation method that resamples feature values from the underlying distribution.
    \item We demonstrate that it is modality agnostic, in the sense that it can be applied to a single modality or multiple heterogeneous modalities. 
    \item We match state-of-the-art performance on the CMU-MOSEI by simultaneously augmenting all modalities.
    \item $\mathrm{SeqAug}$ improves performance  for both recurrent and transformer-based architectures.
    \item We show that $\mathrm{SeqAug}$ requires $20\%$ less training data to match or outperform models trained on the entire dataset. 
\end{enumerate}

\begin{figure*}[!htb]
    \centering
    \begin{subfigure}[h]{.30\textwidth}
        \centering
        \centerline{\scalebox{.33}{\tikzset{every picture/.style={line width=0.75pt}} 

\begin{tikzpicture}[x=0.75pt,y=0.75pt,yscale=-1,xscale=1]

\draw  [color={rgb, 255:red, 0; green, 0; blue, 0 }  ,draw opacity=1 ][fill={rgb, 255:red, 217; green, 249; blue, 194 }  ,fill opacity=1 ] (18,14) -- (68,14) -- (68,64) -- (18,64) -- cycle ;
\draw  [color={rgb, 255:red, 0; green, 0; blue, 0 }  ,draw opacity=1 ][fill={rgb, 255:red, 217; green, 249; blue, 194 }  ,fill opacity=1 ] (18,64) -- (68,64) -- (68,114) -- (18,114) -- cycle ;
\draw  [color={rgb, 255:red, 0; green, 0; blue, 0 }  ,draw opacity=1 ][fill={rgb, 255:red, 217; green, 249; blue, 194 }  ,fill opacity=1 ] (18,114) -- (68,114) -- (68,164) -- (18,164) -- cycle ;
\draw  [color={rgb, 255:red, 0; green, 0; blue, 0 }  ,draw opacity=1 ][fill={rgb, 255:red, 217; green, 249; blue, 194 }  ,fill opacity=1 ] (18,164) -- (68,164) -- (68,214) -- (18,214) -- cycle ;
\draw  [color={rgb, 255:red, 0; green, 0; blue, 0 }  ,draw opacity=1 ][fill={rgb, 255:red, 217; green, 249; blue, 194 }  ,fill opacity=1 ] (18,214) -- (68,214) -- (68,264) -- (18,264) -- cycle ;
\draw  [color={rgb, 255:red, 0; green, 0; blue, 0 }  ,draw opacity=1 ][fill={rgb, 255:red, 203; green, 233; blue, 208 }  ,fill opacity=1 ] (130,14) -- (180,14) -- (180,64) -- (130,64) -- cycle ;
\draw  [color={rgb, 255:red, 0; green, 0; blue, 0 }  ,draw opacity=1 ][fill={rgb, 255:red, 203; green, 233; blue, 208 }  ,fill opacity=1 ] (130,64) -- (180,64) -- (180,114) -- (130,114) -- cycle ;
\draw  [color={rgb, 255:red, 0; green, 0; blue, 0 }  ,draw opacity=1 ][fill={rgb, 255:red, 203; green, 233; blue, 208 }  ,fill opacity=1 ] (130,114) -- (180,114) -- (180,164) -- (130,164) -- cycle ;
\draw  [color={rgb, 255:red, 0; green, 0; blue, 0 }  ,draw opacity=1 ][fill={rgb, 255:red, 203; green, 233; blue, 208 }  ,fill opacity=1 ] (130,164) -- (180,164) -- (180,214) -- (130,214) -- cycle ;
\draw  [color={rgb, 255:red, 0; green, 0; blue, 0 }  ,draw opacity=1 ][fill={rgb, 255:red, 203; green, 233; blue, 208 }  ,fill opacity=1 ] (130,214) -- (180,214) -- (180,264) -- (130,264) -- cycle ;
\draw  [color={rgb, 255:red, 0; green, 0; blue, 0 }  ,draw opacity=1 ][fill={rgb, 255:red, 192; green, 222; blue, 217 }  ,fill opacity=1 ] (241,14) -- (291,14) -- (291,64) -- (241,64) -- cycle ;
\draw  [color={rgb, 255:red, 0; green, 0; blue, 0 }  ,draw opacity=1 ][fill={rgb, 255:red, 192; green, 222; blue, 217 }  ,fill opacity=1 ] (241,64) -- (291,64) -- (291,114) -- (241,114) -- cycle ;
\draw  [color={rgb, 255:red, 0; green, 0; blue, 0 }  ,draw opacity=1 ][fill={rgb, 255:red, 192; green, 222; blue, 217 }  ,fill opacity=1 ] (241,114) -- (291,114) -- (291,164) -- (241,164) -- cycle ;
\draw  [color={rgb, 255:red, 0; green, 0; blue, 0 }  ,draw opacity=1 ][fill={rgb, 255:red, 192; green, 222; blue, 217 }  ,fill opacity=1 ] (241,164) -- (291,164) -- (291,214) -- (241,214) -- cycle ;
\draw  [color={rgb, 255:red, 0; green, 0; blue, 0 }  ,draw opacity=1 ][fill={rgb, 255:red, 192; green, 222; blue, 217 }  ,fill opacity=1 ] (241,214) -- (291,214) -- (291,264) -- (241,264) -- cycle ;
\draw  [color={rgb, 255:red, 0; green, 0; blue, 0 }  ,draw opacity=1 ][fill={rgb, 255:red, 182; green, 213; blue, 225 }  ,fill opacity=1 ] (353,14) -- (403,14) -- (403,64) -- (353,64) -- cycle ;
\draw  [color={rgb, 255:red, 0; green, 0; blue, 0 }  ,draw opacity=1 ][fill={rgb, 255:red, 182; green, 213; blue, 225 }  ,fill opacity=1 ] (353,64) -- (403,64) -- (403,114) -- (353,114) -- cycle ;
\draw  [color={rgb, 255:red, 0; green, 0; blue, 0 }  ,draw opacity=1 ][fill={rgb, 255:red, 182; green, 213; blue, 225 }  ,fill opacity=1 ] (353,114) -- (403,114) -- (403,164) -- (353,164) -- cycle ;
\draw  [color={rgb, 255:red, 0; green, 0; blue, 0 }  ,draw opacity=1 ][fill={rgb, 255:red, 182; green, 213; blue, 225 }  ,fill opacity=1 ] (353,164) -- (403,164) -- (403,214) -- (353,214) -- cycle ;
\draw  [color={rgb, 255:red, 0; green, 0; blue, 0 }  ,draw opacity=1 ][fill={rgb, 255:red, 182; green, 213; blue, 225 }  ,fill opacity=1 ] (353,214) -- (403,214) -- (403,264) -- (353,264) -- cycle ;
\draw  [color={rgb, 255:red, 0; green, 0; blue, 0 }  ,draw opacity=1 ][fill={rgb, 255:red, 168; green, 201; blue, 234 }  ,fill opacity=1 ] (465,14) -- (515,14) -- (515,64) -- (465,64) -- cycle ;
\draw  [color={rgb, 255:red, 0; green, 0; blue, 0 }  ,draw opacity=1 ][fill={rgb, 255:red, 168; green, 201; blue, 234 }  ,fill opacity=1 ] (465,64) -- (515,64) -- (515,114) -- (465,114) -- cycle ;
\draw  [color={rgb, 255:red, 0; green, 0; blue, 0 }  ,draw opacity=1 ][fill={rgb, 255:red, 168; green, 201; blue, 234 }  ,fill opacity=1 ] (465,114) -- (515,114) -- (515,164) -- (465,164) -- cycle ;
\draw  [color={rgb, 255:red, 0; green, 0; blue, 0 }  ,draw opacity=1 ][fill={rgb, 255:red, 168; green, 201; blue, 234 }  ,fill opacity=1 ] (465,164) -- (515,164) -- (515,214) -- (465,214) -- cycle ;
\draw  [color={rgb, 255:red, 0; green, 0; blue, 0 }  ,draw opacity=1 ][fill={rgb, 255:red, 168; green, 201; blue, 234 }  ,fill opacity=1 ] (465,214) -- (515,214) -- (515,264) -- (465,264) -- cycle ;
\draw  [color={rgb, 255:red, 0; green, 0; blue, 0 }  ,draw opacity=1 ][fill={rgb, 255:red, 132; green, 178; blue, 254 }  ,fill opacity=1 ] (577,14) -- (627,14) -- (627,64) -- (577,64) -- cycle ;
\draw  [color={rgb, 255:red, 0; green, 0; blue, 0 }  ,draw opacity=1 ][fill={rgb, 255:red, 132; green, 178; blue, 254 }  ,fill opacity=1 ] (577,64) -- (627,64) -- (627,114) -- (577,114) -- cycle ;
\draw  [color={rgb, 255:red, 0; green, 0; blue, 0 }  ,draw opacity=1 ][fill={rgb, 255:red, 132; green, 178; blue, 254 }  ,fill opacity=1 ] (577,114) -- (627,114) -- (627,164) -- (577,164) -- cycle ;
\draw  [color={rgb, 255:red, 0; green, 0; blue, 0 }  ,draw opacity=1 ][fill={rgb, 255:red, 132; green, 178; blue, 254 }  ,fill opacity=1 ] (577,164) -- (627,164) -- (627,214) -- (577,214) -- cycle ;
\draw  [color={rgb, 255:red, 0; green, 0; blue, 0 }  ,draw opacity=1 ][fill={rgb, 255:red, 132; green, 178; blue, 254 }  ,fill opacity=1 ] (577,214) -- (627,214) -- (627,264) -- (577,264) -- cycle ;
\draw [color={rgb, 255:red, 0; green, 0; blue, 0 }  ,draw opacity=1 ][line width=1.5]    (27,313) -- (617,312.01) ;
\draw [shift={(620,312)}, rotate = 179.9] [color={rgb, 255:red, 0; green, 0; blue, 0 }  ,draw opacity=1 ][line width=1.5]    (14.21,-4.28) .. controls (9.04,-1.82) and (4.3,-0.39) .. (0,0) .. controls (4.3,0.39) and (9.04,1.82) .. (14.21,4.28)   ;

\draw (261,274.4) node [anchor=north west][inner sep=0.75pt]  [font=\huge,color={rgb, 255:red, 0; green, 0; blue, 0 }  ,opacity=1 ]  {$\mathbf{Time\ axis}$};

\end{tikzpicture}}}
        \caption{A sequence of features. Different colors indicate different timesteps. Time axis arrow shows time progression.}
        \label{fig:seq0}
    \end{subfigure}
    \hfill
    \begin{subfigure}[h]{.30\textwidth}
        \centering
        \centerline{\scalebox{.33}{\tikzset{every picture/.style={line width=0.75pt}} 

\begin{tikzpicture}[x=0.75pt,y=0.75pt,yscale=-1,xscale=1]

\draw  [color={rgb, 255:red, 0; green, 0; blue, 0 }  ,draw opacity=1 ][fill={rgb, 255:red, 217; green, 249; blue, 194 }  ,fill opacity=1 ] (18,14) -- (68,14) -- (68,64) -- (18,64) -- cycle ;
\draw  [color={rgb, 255:red, 0; green, 0; blue, 0 }  ,draw opacity=1 ][fill={rgb, 255:red, 217; green, 249; blue, 194 }  ,fill opacity=1 ] (18,64) -- (68,64) -- (68,114) -- (18,114) -- cycle ;
\draw  [color={rgb, 255:red, 0; green, 0; blue, 0 }  ,draw opacity=1 ][fill={rgb, 255:red, 217; green, 249; blue, 194 }  ,fill opacity=1 ] (18,114) -- (68,114) -- (68,164) -- (18,164) -- cycle ;
\draw  [color={rgb, 255:red, 0; green, 0; blue, 0 }  ,draw opacity=1 ][fill={rgb, 255:red, 217; green, 249; blue, 194 }  ,fill opacity=1 ] (18,164) -- (68,164) -- (68,214) -- (18,214) -- cycle ;
\draw  [color={rgb, 255:red, 0; green, 0; blue, 0 }  ,draw opacity=1 ][fill={rgb, 255:red, 217; green, 249; blue, 194 }  ,fill opacity=1 ] (18,214) -- (68,214) -- (68,264) -- (18,264) -- cycle ;
\draw  [color={rgb, 255:red, 0; green, 0; blue, 0 }  ,draw opacity=1 ][fill={rgb, 255:red, 203; green, 233; blue, 208 }  ,fill opacity=1 ] (130,14) -- (180,14) -- (180,64) -- (130,64) -- cycle ;
\draw  [color={rgb, 255:red, 0; green, 0; blue, 0 }  ,draw opacity=1 ][fill={rgb, 255:red, 203; green, 233; blue, 208 }  ,fill opacity=1 ] (130,64) -- (180,64) -- (180,114) -- (130,114) -- cycle ;
\draw  [color={rgb, 255:red, 0; green, 0; blue, 0 }  ,draw opacity=1 ][fill={rgb, 255:red, 203; green, 233; blue, 208 }  ,fill opacity=1 ] (130,114) -- (180,114) -- (180,164) -- (130,164) -- cycle ;
\draw  [color={rgb, 255:red, 0; green, 0; blue, 0 }  ,draw opacity=1 ][fill={rgb, 255:red, 203; green, 233; blue, 208 }  ,fill opacity=1 ] (130,164) -- (180,164) -- (180,214) -- (130,214) -- cycle ;
\draw  [color={rgb, 255:red, 0; green, 0; blue, 0 }  ,draw opacity=1 ][fill={rgb, 255:red, 203; green, 233; blue, 208 }  ,fill opacity=1 ] (130,214) -- (180,214) -- (180,264) -- (130,264) -- cycle ;
\draw  [color={rgb, 255:red, 0; green, 0; blue, 0 }  ,draw opacity=1 ][fill={rgb, 255:red, 192; green, 222; blue, 217 }  ,fill opacity=1 ] (241,14) -- (291,14) -- (291,64) -- (241,64) -- cycle ;
\draw  [color={rgb, 255:red, 0; green, 0; blue, 0 }  ,draw opacity=1 ][fill={rgb, 255:red, 192; green, 222; blue, 217 }  ,fill opacity=1 ] (241,64) -- (291,64) -- (291,114) -- (241,114) -- cycle ;
\draw  [color={rgb, 255:red, 0; green, 0; blue, 0 }  ,draw opacity=1 ][fill={rgb, 255:red, 192; green, 222; blue, 217 }  ,fill opacity=1 ] (241,164) -- (291,164) -- (291,214) -- (241,214) -- cycle ;
\draw  [color={rgb, 255:red, 0; green, 0; blue, 0 }  ,draw opacity=1 ][fill={rgb, 255:red, 192; green, 222; blue, 217 }  ,fill opacity=1 ] (241,214) -- (291,214) -- (291,264) -- (241,264) -- cycle ;
\draw  [color={rgb, 255:red, 0; green, 0; blue, 0 }  ,draw opacity=1 ][fill={rgb, 255:red, 182; green, 213; blue, 225 }  ,fill opacity=1 ] (353,14) -- (403,14) -- (403,64) -- (353,64) -- cycle ;
\draw  [color={rgb, 255:red, 0; green, 0; blue, 0 }  ,draw opacity=1 ][fill={rgb, 255:red, 182; green, 213; blue, 225 }  ,fill opacity=1 ] (353,64) -- (403,64) -- (403,114) -- (353,114) -- cycle ;
\draw  [color={rgb, 255:red, 0; green, 0; blue, 0 }  ,draw opacity=1 ][fill={rgb, 255:red, 182; green, 213; blue, 225 }  ,fill opacity=1 ] (353,164) -- (403,164) -- (403,214) -- (353,214) -- cycle ;
\draw  [color={rgb, 255:red, 0; green, 0; blue, 0 }  ,draw opacity=1 ][fill={rgb, 255:red, 182; green, 213; blue, 225 }  ,fill opacity=1 ] (353,214) -- (403,214) -- (403,264) -- (353,264) -- cycle ;
\draw  [color={rgb, 255:red, 0; green, 0; blue, 0 }  ,draw opacity=1 ][fill={rgb, 255:red, 168; green, 201; blue, 234 }  ,fill opacity=1 ] (465,14) -- (515,14) -- (515,64) -- (465,64) -- cycle ;
\draw  [color={rgb, 255:red, 0; green, 0; blue, 0 }  ,draw opacity=1 ][fill={rgb, 255:red, 168; green, 201; blue, 234 }  ,fill opacity=1 ] (465,64) -- (515,64) -- (515,114) -- (465,114) -- cycle ;
\draw  [color={rgb, 255:red, 0; green, 0; blue, 0 }  ,draw opacity=1 ][fill={rgb, 255:red, 168; green, 201; blue, 234 }  ,fill opacity=1 ] (465,164) -- (515,164) -- (515,214) -- (465,214) -- cycle ;
\draw  [color={rgb, 255:red, 0; green, 0; blue, 0 }  ,draw opacity=1 ][fill={rgb, 255:red, 168; green, 201; blue, 234 }  ,fill opacity=1 ] (465,214) -- (515,214) -- (515,264) -- (465,264) -- cycle ;
\draw  [color={rgb, 255:red, 0; green, 0; blue, 0 }  ,draw opacity=1 ][fill={rgb, 255:red, 132; green, 178; blue, 254 }  ,fill opacity=1 ] (577,14) -- (627,14) -- (627,64) -- (577,64) -- cycle ;
\draw  [color={rgb, 255:red, 0; green, 0; blue, 0 }  ,draw opacity=1 ][fill={rgb, 255:red, 132; green, 178; blue, 254 }  ,fill opacity=1 ] (577,64) -- (627,64) -- (627,114) -- (577,114) -- cycle ;
\draw  [color={rgb, 255:red, 0; green, 0; blue, 0 }  ,draw opacity=1 ][fill={rgb, 255:red, 132; green, 178; blue, 254 }  ,fill opacity=1 ] (577,164) -- (627,164) -- (627,214) -- (577,214) -- cycle ;
\draw  [color={rgb, 255:red, 0; green, 0; blue, 0 }  ,draw opacity=1 ][fill={rgb, 255:red, 132; green, 178; blue, 254 }  ,fill opacity=1 ] (577,214) -- (627,214) -- (627,264) -- (577,264) -- cycle ;
\draw [color={rgb, 255:red, 0; green, 0; blue, 0 }  ,draw opacity=1 ][line width=1.5]    (27,313) -- (617,312.01) ;
\draw [shift={(620,312)}, rotate = 179.9] [color={rgb, 255:red, 0; green, 0; blue, 0 }  ,draw opacity=1 ][line width=1.5]    (14.21,-4.28) .. controls (9.04,-1.82) and (4.3,-0.39) .. (0,0) .. controls (4.3,0.39) and (9.04,1.82) .. (14.21,4.28)   ;
\draw  [color={rgb, 255:red, 0; green, 0; blue, 0 }  ,draw opacity=1 ][fill={rgb, 255:red, 203; green, 233; blue, 208 }  ,fill opacity=1 ] (130,114) -- (180,114) -- (180,164) -- (130,164) -- cycle ;
\draw  [color={rgb, 255:red, 0; green, 0; blue, 0 }  ,draw opacity=1 ][fill={rgb, 255:red, 203; green, 233; blue, 208 }  ,fill opacity=1 ] (241,114) -- (291,114) -- (291,164) -- (241,164) -- cycle ;
\draw  [color={rgb, 255:red, 0; green, 0; blue, 0 }  ,draw opacity=1 ][fill={rgb, 255:red, 192; green, 222; blue, 217 }  ,fill opacity=1 ] (241,114) -- (291,114) -- (291,164) -- (241,164) -- cycle ;
\draw  [color={rgb, 255:red, 0; green, 0; blue, 0 }  ,draw opacity=1 ][fill={rgb, 255:red, 182; green, 213; blue, 225 }  ,fill opacity=1 ] (353,114) -- (403,114) -- (403,164) -- (353,164) -- cycle ;
\draw  [color={rgb, 255:red, 0; green, 0; blue, 0 }  ,draw opacity=1 ][fill={rgb, 255:red, 203; green, 233; blue, 208 }  ,fill opacity=1 ] (465,114) -- (515,114) -- (515,164) -- (465,164) -- cycle ;
\draw  [color={rgb, 255:red, 0; green, 0; blue, 0 }  ,draw opacity=1 ][fill={rgb, 255:red, 168; green, 201; blue, 234 }  ,fill opacity=1 ] (465,114) -- (515,114) -- (515,164) -- (465,164) -- cycle ;
\draw  [color={rgb, 255:red, 0; green, 0; blue, 0 }  ,draw opacity=1 ][fill={rgb, 255:red, 203; green, 233; blue, 208 }  ,fill opacity=1 ] (577,114) -- (627,114) -- (627,164) -- (577,164) -- cycle ;
\draw  [color={rgb, 255:red, 0; green, 0; blue, 0 }  ,draw opacity=1 ][fill={rgb, 255:red, 132; green, 178; blue, 254 }  ,fill opacity=1 ] (577,114) -- (627,114) -- (627,164) -- (577,164) -- cycle ;
\draw  [color={rgb, 255:red, 241; green, 26; blue, 10 }  ,draw opacity=1 ][line width=3.75]  (8,125) .. controls (8,119.48) and (12.48,115) .. (18,115) -- (632,115) .. controls (637.52,115) and (642,119.48) .. (642,125) -- (642,155) .. controls (642,160.52) and (637.52,165) .. (632,165) -- (18,165) .. controls (12.48,165) and (8,160.52) .. (8,155) -- cycle ;

\draw (261,274.4) node [anchor=north west][inner sep=0.75pt]  [font=\huge,color={rgb, 255:red, 0; green, 0; blue, 0 }  ,opacity=1 ]  {$\mathbf{Time\ axis}$};

\end{tikzpicture}}}
        \caption{\textbf{Feature Address Sampling}: We randomly pick some feature dimensions across all timesteps (red rectangle).}
        \label{fig:seq1}
    \end{subfigure}
    \hfill
    \begin{subfigure}[h]{.30\textwidth}
        \centering
        \centerline{\scalebox{.33}{\input{mmaug_2.tikz}}}
        \caption{\textbf{Time Resampling}: We resample each feature value from the sequence itself. Can be modeled as permutation.}
        \label{fig:seq2}
    \end{subfigure}
    \caption{\SeqAug{} approach.}
    \label{fig:MMAug}
\end{figure*}

\section{Feature Distribution Resampling} \label{sec:format}
We present a novel feature augmentation technique called $\mathrm{SeqAug}$, which operates on sequential data and can be applied in a modality agnostic manner. It is compatible with a wide range of feature types, such as GloVe embeddings.

\subsection{SeqAug}
Assume a sequence of extracted features $\bigl\{x_t\bigr\}_{t=0}^T$, where $x_t$ lies in $\mathbb{R}^d$ and $d$ is the feature dimensionality (see Fig.~\ref{fig:seq0}). During this work we assume that any instance of the sequence may be sampled from an underlying distribution as in Fig.~\ref{fig:p_feat}. From a mathematical perspective we have discarded the time-varying nature of the extracted features. Specifically we assume that all features may be drawn from some unknown distribution as of 
\begin{equation}
 x_t \sim p_{feat}(x|c)   
\end{equation}
where $c$ denotes the class label \footnote{This holds for the dataset we examine. However in general it should be conditioned on the utterance and not the class label.}. In the multimodal scenario we sample each stream from its unimodal distribution which is modeled as the marginal distribution of the joint multimodal one. Now consider all the available feature indices defined as $\mathcal{D}=\{1,\cdots,d\}$. We call each index lying in $\mathcal{D}$ a feature address. We describe the proposed approach in two steps. 

\noindent
\textbf{Feature Address Sampling} Given a sequence  $\bigl\{x_t\bigr\}_{t=0}^T$, we randomly pick some feature addresses from the set $\mathcal{D}$. These addresses are preserved across all timesteps within the sequence (see red reactangle in Fig.~\ref{fig:seq1}).

\noindent
\textbf{Time Resampling} We perform a random permutation $\pi$ of the ordered time indices. The feature addresses sampled in the previous step are permuted along the time-axis according to $\pi$ (see Fig.~\ref{fig:seq2}). The rest feature addresses remain the same. 

For example in Fig.~\ref{fig:MMAug}, we have randomly picked only feature address set ${2}$ (could have been be any other subset of $\mathcal{D}$) and $\pi=(2,5,4,1,0,3)$. This means that the second feature value is permuted across the time-axis according to $\pi$.


\subsection{Implementation Details}
\noindent{\textbf{Design Choices}} \SeqAug{} takes a sequence of features and performs two basic actions (steps). In \textit{Feature Address Sampling} Step, it randomly sets the percentage of addresses that are going to be permuted, e.g., $35\%$, and then randomly picks them. In \textit{Time Resampling} Step it permutes these features across the time-axis. This process is repeated for each modality. 

In an attempt to make the method more robust to hyperparameter choice, we introduce a probability distribution $\mathcal{P}$ from which we sample the number of addresses which are going to be sampled. We find that beta distribution, $\mathcal{B}(\alpha, \alpha)$, is suited for transformer-based models while folded normal for recurrent ones. Furthermore, we follow~\cite{zoneout} and preserve the sampled feature addresses across timesteps of the sequence.

\noindent{\textbf{Complexity}} The computational complexity of \SeqAug{} during training can be expressed as $\mathcal{O}(ML)$ for $M$ independent sequences of features, each with a length of $L$ and dimensionality of $d$, under the assumption of constant time in element-wise operations. Specifically, in the scenario where $M$ is equal to $1$, \SeqAug{} scales linearly with respect to the sequence length. It is important to note that \SeqAug{} is deactivated during inference.

\noindent{\textbf{Hyperparameters}} To introduce a minimum set of hyperparameters, we tie the beta distribution hyperparameters and tune a single $\alpha \in [0.0, 1.0]$. As for the folded-normal, we fix the deviation $\sigma$ in some small value, i.e., $0.01$, and only tune the mean $\mu \in [0, 1]$. We use an additional (independent) hyperparameter when more than one modalities are involved.

\section{Experimental Setup} \label{sec:exp-setup}

\subsection{Dataset}
To establish 
the proposed \SeqAug{}, we present experimental results on the CMU-MOSEI sentiment analysis dataset \cite{mosei}.
In particular, the dataset contains $23,454$ YouTube video clips of movie reviews, accompanied by annotated sentiment scores from -3 (strongly negative) to 3 (strongly positive). Our approach to sentiment analysis is in line with the literature, where we treat it as a regression problem \cite{mao-etal-2022-sena}.

\subsection{Feature Set}
To ensure that our approach is comparable with existing literature, we adopt the same feature sets proposed in the original MOSEI paper \cite{mosei}. Specifically, audio sequences are encoded using COVAREP features \cite{degottex2014covarep}, visual sequences are represented using Facet features\footnote{https://imotions.com/guides/facial-expression-analysis/}, and transcriptions are segmented into words and represented using GloVe \cite{pennington-etal-2014-glove}. All sequences are word-aligned using P2FA \cite{yuan2008speaker}. To facilitate fair comparisons, we adhere to the standard train/val/test splits provided. We denote as $Acc^2$ and $Acc^7$ the binary and seven class accuracies respectively, while MAE indicates the Mean Absolute Error.

\subsection{Architectures}
We employ two models of distinct architectural flavors, namely a multimodal transformer~\cite{tsai-etal-2019-multimodal} (MulT) and the \textit{Baseline} (shown in Fig~1 of~\cite{georgiou21_interspeech}) recurrent-based architecture (MMRnn) described in~\cite{georgiou21_interspeech, mmlatch}.
We follow the exact setting, hyperparameter choice, and optimization setup described in the original papers. We follow their official available implementations\footnote{https://github.com/yaohungt/Multimodal-Transformer} \footnote{ https://github.com/efthymisgeo/multimodal-masking}.

For MulT, we use a beta distribution for \SeqAug{} and tune it in the range $[0.1, 1.0]$ with step $0.2$. For MMRnn, we found that folded-normal is more robust and effective. The mean $\mu$ is tuned in $[0.1, 0.5]$ with step $0.05$ for each modality independently.
All experiments were run on a single NVIDIA GeForce RTX 2080 Ti (11GB).

\section{Experiments} \label{sec:experiments}

\subsection{Modality agnostic feature augmentation}
In this experiment, we assess the impact of the proposed feature-augmentation at each modality independently. We choose a multimodal architecture (MMRnn) and apply \SeqAug{} to each modality separately. The mean of the folded-normal distribution is tuned for each modality, and the best value is reported. The results indicate that the text-augmented (T) model performs better with a mean of $0.15$ while for the audio (A) and visual-augmented (V) models $0.4$ and $0.35$ worked better. The results are presented in Table~\ref{tab:unimodal}.

The baseline (MMRnn) is outperformed by all unimodal-augmented models, indicating that the proposed method is suitable for multiple modalities, making it modality agnostic. Interestingly, no single modality consistently outperforms all others in terms of the examined metrics, suggesting that each modality contributes in a different manner to the overall model performance. For instance, the text-augmented (T) model performs better in terms of both accuracies, while the visual-augmented (V) model has the lowest mean absolute error (MAE). This result motivated the authors to augment all modalities at once. 

\begin{table}[htb!]
  \begin{center}
  \caption{\SeqAug{} applied to individual modalities. T, A, V represent text, audio, and visual augmented models, respectively. The reported results are an average of five independent runs, with the best performing scores underlined.}
  \resizebox{0.83\columnwidth}{!}{
    \begin{tabular}{lccccccccr}
      \multicolumn{5}{c}{\textbf{Metrics}}
        & MMRnn$^*$
        & \textsc{T}
        & A
        & V \\
      \midrule 
      \rowcolor{Graey}
      \multicolumn{5}{c}{Acc$2$ $(\%)$} & $81.90$ & $\underline{82.22}$ & $82.15$ & $82.03$ \\
      \midrule 
      \rowcolor{Grey}
      \multicolumn{5}{c}{Acc$7$ $(\%)$} & $51.23$ & $\underline{51.72}$ & $51.57$ & $51.68$\\
      \midrule
      \rowcolor{Gray}
      \multicolumn{5}{c}{MAE} & $59.23$ & $58.99$  & $58.73$ & $\underline{58.69}$\\
      \bottomrule
    \end{tabular}
    }
    \label{tab:unimodal}
  \end{center}
\end{table}

\begin{table}[htb!]
  \begin{center}
  \caption{Evaluation of \SeqAug{} on multimodal architectures. The reproduced architecture is denoted by $^*$, while \texttt{+}\SeqAug{} represents the augmented architecture. The $\Delta$ rows depict the improvement for each metric (for MAE lower is better).}
  \resizebox{0.99\columnwidth}{!}{
    \begin{tabular}{lccccccr}
      \multicolumn{4}{c}{\textbf{Model}}
        & Acc$2$ $(\%)$
        & Acc$7$ $(\%)$
        & MAE \\
      \midrule 
        \multicolumn{4}{c}{MMRnn$^*$} & $81.90 \pm 0.30$ & $51.23 \pm 0.49$ & $59.23 \pm 0.50$ \\
      \multicolumn{4}{c}{$+$\SeqAug} & $82.48 \pm 0.38$ & $51.90 \pm 0.49$ & $58.50 \pm 0.38$\\
      \rowcolor{Gray}
      \multicolumn{4}{c}{$\Delta$} & \contour{green}{$\uparrow$} $\; 0.58$ & \contour{green}{$\uparrow$} $\; 0.67$ & \contour{green}{$\downarrow$} $\; 0.73$\\
      \toprule
    \multicolumn{4}{c}{MulT$^*$} & $81.05 \pm 0.32$ & $49.98 \pm 0.13$ & $60.80 \pm 0.20$\\
      \multicolumn{4}{c}{$+$\SeqAug{}} & $81.89 \pm 0.17$ &$50.42 \pm 0.36$ & $59.98 \pm 0.23$ \\
      \rowcolor{Gray}
      \multicolumn{4}{c}{$\Delta$} & \contour{green}{$\uparrow$} $\; 0.84$ & \contour{green}{$\uparrow$} $\; 0.44$ & \contour{green}{$\downarrow$} $\; 0.82$\\
      
      \bottomrule
    \end{tabular}
    }
    \label{tab:mmaug_architectures}
  \end{center}
\end{table}

\begin{figure*}[!htb]
    \centering
    \begin{subfigure}[h]{.46\textwidth}
        \centering
        \includegraphics[width=\textwidth]{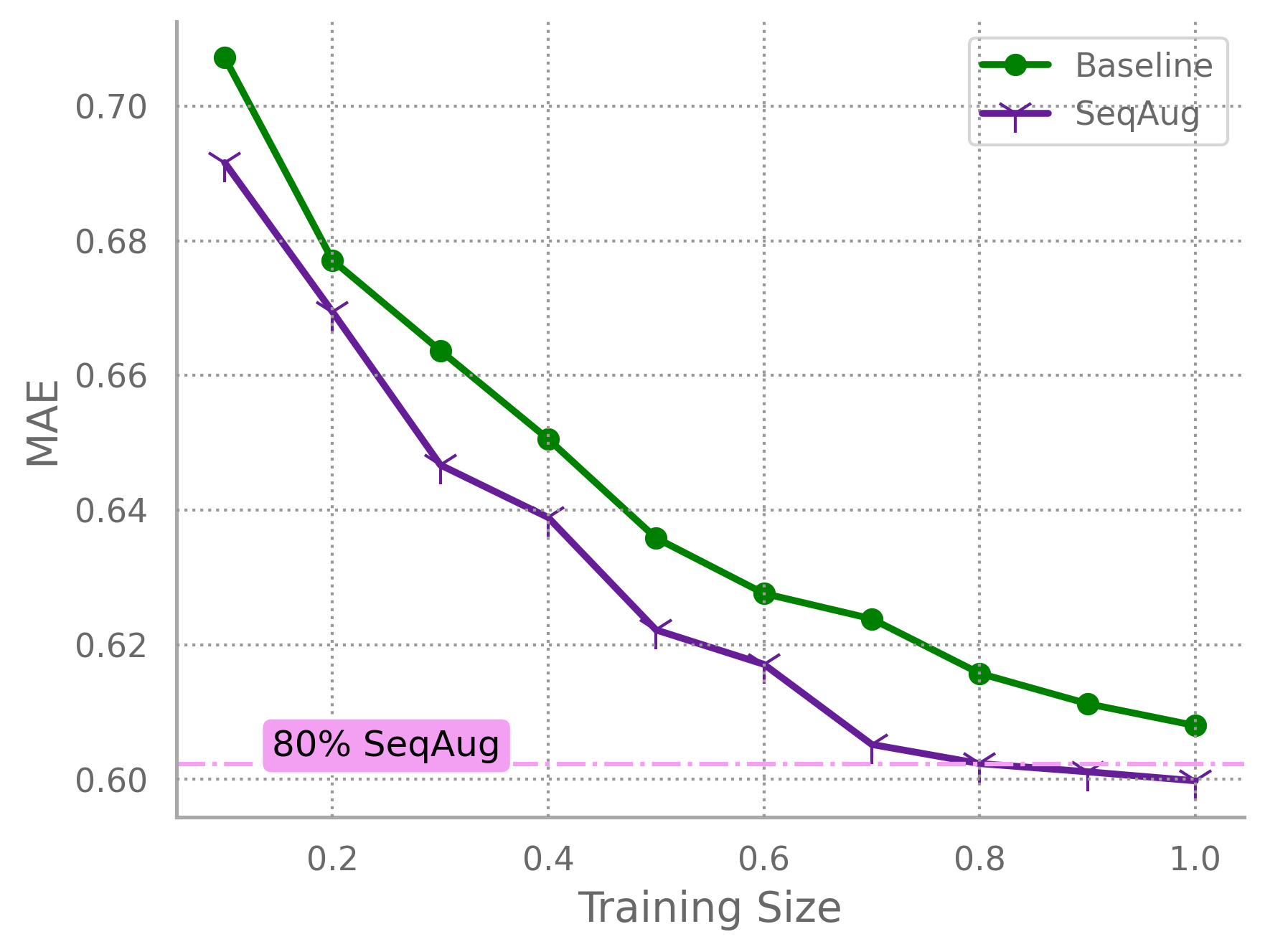}
        \caption{MAE values for training with (purple) and without (green) ($\mathrm{SeqAug}$) across varying training sizes.}
        \label{fig:mae_vs_size}
    \end{subfigure}
    \hfill
    \begin{subfigure}[h]{.46\textwidth}
        \centering
        \includegraphics[width=\textwidth]{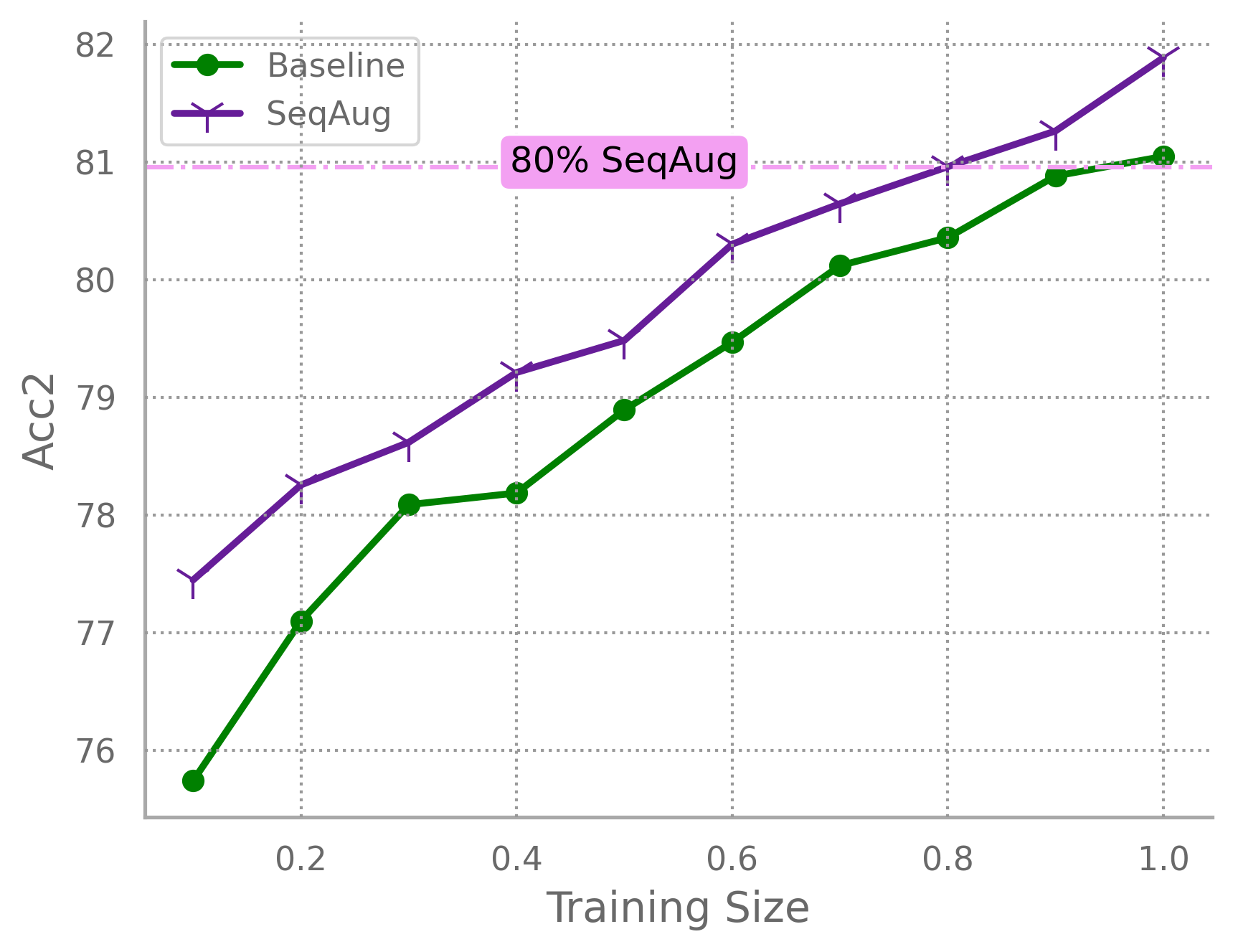}
        \caption{$Acc^2$ scores for training with (purple) and without (green) ($\mathrm{SeqAug}$) across varying training sizes.}
        \label{fig:acc2_vs_size}
    \end{subfigure}
    \caption{$\mathrm{SeqAug}$ in a limited data training setup.}
    \label{fig:MMAug_vs_data}
\end{figure*}

\subsection{Multimodal SeqAug}
To investigate the effectiveness of \SeqAug{} at all three modalities at once, we choose 2 multimodal baselines. The transformer based  MulT~\cite{tsai-etal-2019-multimodal}, and the LSTM-based~\cite{lstm} MMRnn ~\cite{georgiou21_interspeech, mmlatch}. To ensure a fair comparison, we have reproduced these 2 baselines based on their source code and applied \SeqAug{} at both architectures, marked as $+$\SeqAug. 
We report average results over 5 independent runs for all 4 models (Tab.~\ref{tab:mmaug_architectures}).

For augmented MulT (shown as $+$\SeqAug), we start the tuning procedure from default values $1.0$, which works well. We then perform grid search for each parameter and find as best performing triplet $(\alpha_T, \alpha_A, \alpha_V) = (1.0, 0.1, 1.0)$. For augmented MMRnn we use $0.1$ as default values and we perform grid search with $0.05$ step to find as optimal mean values $(\mu_T, \mu_A, \mu_V) = (0.15, 0.1, 0.2)$.

Integrating \SeqAug{} boosts performance consistently across all metrics (average values) for both cases. Furthermore, we observe that the multimodal version of \SeqAug{} achieves higher performance gains compared to the unimodal one (see Tab.~\ref{tab:unimodal}), which admits the benefits of the modality-agnostic nature of \SeqAug{}. Interestingly, we also find that \SeqAug{} is easier to tune in the transformer architecture compared to the recurrent one. Moreover, MulT can handle a more aggressive augmentation scheme due to the beta distribution sampling, which has a mean equal to $0.5$. On the other hand, MMRnn requires a smoother (folded-normal) augmentation to work properly. We attribute this difference to the autoregressive nature of the LSTM.

\vspace{3pt}
\noindent
\textbf{SOTA Discussion} 
Regarding state-of-the-art performance, we observe in the literature that there is no clear winner for the feature set used. The most successful models include MulT~\cite{tsai-etal-2019-multimodal}, M3~\cite{georgiou21_interspeech}, and MMLatch~\cite{mmlatch}. Our experiments show that when MMRnn is combined with \SeqAug{}, we are able to achieve comparable results to the state-of-the-art. Specifically, our model achieves $82.48$ in $Acc^2$, while the highest-performing models (MulT/M3) achieve $82.50$. For $Acc^7$, our model achieves an accuracy of $51.90$, compared to MMLatch's accuracy of $52.00$, and for MAE, our model achieves a value of $0.585$, while MulT's reported value is $0.580$.

We note that while we compare our results with the original MulT paper, we were unable to reproduce their reported results (see MulT* in Tab.\ref{tab:mmaug_architectures}) using their source code [14, 12]. Furthermore, we emphasize that the top-performing models in the literature employ BERT embeddings\cite{hu-etal-2022-unimse}. However, in this study we use the feature set proposed in ~\cite{mosei}, as the main objective of our paper is to introduce and evaluate \SeqAug{}.

\subsection{Limited Training size experiment} \label{sec:training_size}
This experiment tracks the performance gains of \SeqAug{} in relation to the size of the training data, with the aim of demonstrating its effectiveness as a data augmentation method. To this end, we train five MulT models using progressively larger percentages of the training data. The results, as illustrated in Fig.~\ref{fig:MMAug_vs_data}, reveal consistent performance gains/gaps across all training sizes, indicating that \SeqAug{} behaves in a manner consistent with other data augmentation methods and produces useful samples for the task at hand.

Specifically, Fig.~\ref{fig:mae_vs_size} demonstrates that at 80\% of the training data, the \SeqAug{} curve outperforms the baseline model trained with the complete dataset, with a MAE of 0.602 compared to 0.608. This finding suggests that \SeqAug{} produces semantically similar views of the original features, allowing for comparable performance with significantly less data. This conclusion is further supported by the observation that \SeqAug{} requires 20\% fewer data to match the baseline model's binary accuracy (Acc2), achieving a score of 80.98 compared to 81.05 for the baseline model.

\section{Related Work} \label{sec:realted_work}
Conceptually \SeqAug{} is similar to works which aim to sample the underlying data distribution. This may be achieved using a plethora of generative setups such as VAEs~\cite{Kingma2013AutoEncodingVB, multimodal-vae}, GANs~\cite{goodfellowgenerative, Pandeva2019MMGANGA} or normalizing flows~\cite{kingma2018glow}. These generative frameworks have been used in tasks such as semantic segmentation ~\cite{synthetic_semantic_segmentation} and speech emotion recognition ~\cite{chatziagapi19_interspeech}. Our work shares the idea of sampling from the underlying distribution, however it does not involve any additional training or generative network.

From an implementation perspective, a few works share some common ground with \SeqAug{}. Dropout ~\cite{dropout}, has been used as embedding augmentation in many NLP and multimodal scenarios~\cite{tsai-etal-2019-multimodal}. Moreover, CutMix~\cite{cutmix} and MixUp~\cite{mixup} are related to the proposed method in the sense that they modify existing images according to other images in the dataset, i.e., they sample from an unknown pixel distribution. However, both methods are modality specific, do not operate on sequences and also violate the semantic constraint we have set in this work.

ZoneOut~\cite{zoneout} on the other hand, is a work tailored towards recurrent network regularization. It can be seen as a special case of this work since it preserves feature values for all timesteps, which under our formulation can be viewed as resampling a specific value for some feature addresses. ZoneOut however, is limited to recurrent architectures and is also applied at intermediate hidden states rather than the input itself.

\section{Conclusions and Future Work} \label{sec:page}
This study introduces \SeqAug{}, a novel modality-agnostic sequence feature augmentation technique that resamples feature values from the underlying distribution. By replacing existing feature values with a random permutation of the sequence, the method samples the underlying distribution and preserves the semantics of the original sequence. \SeqAug{} is capable of augmenting diverse modalities and shows that augmenting all modalities together and training the multimodal model yields better performance gains than augmenting a single modality. The empirical findings reveal that a more aggressive variant of \SeqAug{} is well-suited for transformer-based architectures, whereas a smoother variant is more suitable for recurrent models. Furthermore, the experimental results indicate that \SeqAug{} attains competitive performance with the current state-of-the-art models across the examined evaluation metrics. Using \SeqAug{} reduces the need for data by 20\% while achieving comparable or even better performance than a model trained on the complete dataset, indicating its efficacy as a feature augmentation method with minimal computational cost.

As future work, we aim to incorporate \SeqAug{} into a broader range of models, including M3, and explore its performance in more diverse modalities and feature sets. Additionally, we intend to compare \SeqAug{} with modality-specific DA methods and establish it as a versatile, plug-and-play augmentation technique. We believe that exploring modality-agnostic transformations is crucial for developing more effective deep learning models.

\bibliographystyle{IEEEtran}
\bibliography{arxiv}

\end{document}